# Evolutionary Discovery of Sequence Acceleration Methods for Slab Geometry Neutron Transport

Japan K. Patel,[a]* Barry D. Ganapol,[b] Anthony Magliari,[c] Matthew C. Schmidt,[a,d] and Todd A. Wareing[c]

[a]*Gateway Scripts, St. Louis, MO*

[b]*University of Arizona, Department of Aerospace and Mechanical Engineering, Tucson, AZ*

[c]*Varian Medical Systems, Palo Alto, CA*

[d]*Washington University in St. Louis, Department of Radiation Oncology, St. Louis, MO*

*E-mail: jpatel@gatewayscripts.com

# Evolutionary Discovery of Sequence Acceleration Methods for Slab Geometry Neutron Transport

We present a genetic programming approach to automatically discover convergence acceleration methods for discrete ordinates solutions of neutron transport problems in slab geometry. Classical acceleration methods such as Aitken's delta-squared and Wynn epsilon assume specific convergence patterns and do not generalize well to the broad set of transport problems encountered in practice. We evolved mathematical formulas specifically tailored to $S_N$ convergence characteristics in this work. The discovered accelerator, featuring second differences and cross-product terms achieved over 75% success rate in improving convergence compared to raw sequences – almost double that observed for classical techniques for the problem set considered. This work demonstrates the potential for discovering novel numerical methods in computational physics via genetic programming and attempts to honor Prof. Ganapol's legacy of advancing experimental mathematics applied to neutron transport.

Keywords: neutron transport; convergence acceleration; genetic programming; discrete ordinates; sequence acceleration

## I. INTRODUCTION

Prof. Ganapol has spent decades developing precise benchmark solutions to radiation transport problems. These benchmarks provide the standards against which production transport codes are verified, ensuring reliability of calculations for reactor design, radiation shielding, and medical physics [1]. In early works, researchers could often compute only a limited number of terms in an infinite series before they hit computational constraints or round-off errors destroyed solutions. Therefore, a naïve approach of simply adding more terms proved to be both expensive and potentially ineffective. Convergence acceleration on the other hand offers a different approach – rather than using brute force, one applies mathematical methods to mine for higher accuracy from less accurate solutions [1]. This approach transforms slowly converging sequences into more rapidly converging ones to obtain desired precision, within the bounds of computational constraints. Prof. Ganapol has been a leading proponent of this acceleration philosophy.

His benchmark compilation for the Nuclear Energy Agency [1] presents solutions obtained through sophisticated techniques like Fourier and Laplace transforms, singular eigenfunction expansions, and the $F_N$ method. To achieve "benchmark quality", typically five digits or more of precision, Prof. Ganapol employed several convergence acceleration techniques including extrapolation methods, and nonlinear approaches [1]. The Wynn-epsilon algorithm [2], which he calls the most elegant of all convergence acceleration methods [1], plays a significant role in accelerating solution sequences. The Euler-Knopp transformation enables efficient evaluation of alternating series [3], while Richardson extrapolation [4] systematically eliminates leading-order error terms. These methods have been applied across diverse problems: radiative transfer [5], nuclear criticality [6], point kinetics [7], the Fokker-Planck equation [8], and boundary layer flows [9]. Our collaborative work with Ganapol on fast burst reactor benchmarks [10], response matrix methods [11, 12], and double $P_N$ method [13] continues this tradition.

Classical acceleration methods make specific assumptions about how sequences behave. Aitken's delta-squared method [14], one of the oldest and most widely used accelerators, assumes geometric convergence and needs three consecutive terms. The Wynn-epsilon algorithm builds an epsilon table through cross-rule recursion and works very well for alternating or oscillating sequences [2]. The Richardson extrapolation exploits known error expansion structures to systematically cancel leading-order error terms [4]. While each method has proven to be highly valuable for several applications in computational physics, their effectiveness depends on whether the relevant sequence conforms to the underlying assumptions. Therefore, in practice, classical accelerators show application-dependent performance. In transport calculations, convergence behavior depends on optical thickness, scattering ratio, and geometric complexity. A method that works for optically thin problems may or may not perform as desired for optically thick ones. Our experiments with $S_N$ solution sequences spanning scattering ratios from 0.001 to 0.999 revealed that both Wynn-epsilon and Aitken's delta-squared method only truly accelerated convergence (with respect to raw sequence) less than 40% of the time. The success rate suggests that assumptions that classical accelerators make do not align well with diverse convergence patterns that $S_N$ solutions exhibit. This naturally leads us to the question we are trying to answer through this paper – can we discover acceleration formulas specifically adapted to transport problems? Rather than applying general-purpose accelerators derived from mathematical abstractions, can we evolve problem-specific formulas optimized for discrete ordinates sequences? Such an approach would align with Prof. Ganapol's philosophy of experimental mathematics – using computational exploration to discover mathematical relationships that might otherwise remain undiscovered.

Genetic programming (GP) [15] offers a framework for this kind of exploration. Unlike traditional optimization which adjusts parameters within a fixed functional form, GP evolves the mathematical structure itself by mechanisms inspired by biological evolution. A population of candidate formulas, represented as expression trees, undergoes selection, crossover, and mutation over successive generations. Formulas that best satisfy a fitness criterion are allowed to reproduce, gradually steering the population towards improved solutions. This approach has been shown to be successful, most notably by Schmidt and Lipson [16], which used symbolic regression to rediscover laws of physics from experimental data without prior knowledge of the underlying physics.

Applying genetic programming to discover convergence acceleration formulas represents a novel intersection of evolutionary computation and numerical analysis. By training on sequences of $S_N$ solutions spanning a diverse set of physical parameters, GP can identify mathematical structures that represent patterns specific to transport calculations. The resulting formulas emerge directly from data and therefore can potentially capture relationships that "trained" humans might overlook. In this paper, we demonstrate that GP can discover effective convergence acceleration formulas for $S_N$ calculations. For the test data considered in this study, GP-evolved formulas were observed to achieve approximately 78% success rate, almost double the performance of classical methods. The discovered formula exhibits structural features like second differences and cross-terms that appear to be adapted to underlying $S_N$ convergence patterns.

The remainder of this paper is organized as follows. Section II introduces symbolic regression and genetic programming fundamentals, including terminal and function sets, tree representation, and evolution dynamics. Section III details the application to transport problems – dataset generation, GP configuration, and computational setup. Section IV presents the discovered formula and discusses observations. Section V concludes this paper with summary and outlook.

## II. BACKGROUND: SYMBOLIC REGRESSION AND GENETIC PROGRAMMING

The goal of this work is to discover mathematical formulas that accelerate convergence of $S_N$ sequences. Rather than assuming a functional form and fitting parameters, we want to discover the structure itself to emerge from data. Symbolic regression searches for both – the parameter set and the underlying mathematical structure simultaneously. Therefore, the output is not just optimized coefficients; it returns an entire mathematical expression that is appropriate for the underlying data. This distinction matters for discovering convergence acceleration formulas because we do not know, a priori, what mathematical form is needed to represent the underlying accelerator structure. Classical accelerators like Aitken's delta-squared method emerged from theoretical analysis of geometric sequences. However, transport sequences may require different structures entirely. Symbolic regression allows the data to guide us toward appropriate functional forms without imposing restrictive assumptions. The challenge is that the space of possible mathematical expressions is large – even with a small set of operators and variables, the number of valid expressions grows combinatorially with the complexity. GP provides a way to navigate this space efficiently.

### II.A. Genetic Programming Preliminaries

Genetic programming presents mathematical expressions as tree structures [15]. Each tree encodes a formula that can be evaluated, compared against others, and modified through evolutionary operators. The building blocks of GP trees fall into two categories – terminals and functions. Terminals are the leaf nodes – variables or constants that constitute the function. They take no arguments. Functions are internal nodes. These are operators that combine related nodes below them. Together, the terminal set and the function set define the dictionary from which GP can construct relevant relations. The choice of these sets is problem-dependent and determines what kind of formulas can be discovered.

A GP tree is evaluated recursively from the leaves leftward (or upward depending on tree orientation). Each terminal returns its value, and each function applies its operation to the values coming from the nodes below them. The root node produces the final result. The tree depth – the longest path from the root to leaf – is typically constrained to control formula complexity. For GP to work reliably, every possible tree must produce a valid numerical output. This property is called closure, and it is enforced by setting failing outputs, e.g. when the denominator is near zero, to a default value.

To illustrate trees, consider Aitken's classical method, where given a sequence $S_0, S_1, S_2\ldots$ is accelerated according to:

$$A_n = S_n - \frac{(S_n - S_{n-1})^2}{S_n - 2S_{n-1} + S_{n-2}}. \tag{1}$$

Figure 1 shows its tree representation. The root is a subtraction node [-]. Its bottom branch leads to the terminal $S_n$. Its top branch leads to a division node [÷], with the squared first difference as numerator and the second difference as the denominator. Each path from the root to the leaf traces a sequence of operations. This tree has depth 4. If we were to evolve acceleration formulas, trees like this would compete, combine, and mutate to potentially return structures distinct from this classical form. The "tree" here differs from decision trees used in methods like random forests [17]. A decision tree whether for classification or regression splits data based on feature thresholds. Each internal node asks, "is feature X greater than value Y?" and routes accordingly. A random forest averages predictions from many such trees. For regression tasks, the output is a numerical prediction, not a formula. This is essentially just a procedure that maps inputs to outputs. A GP tree, in contrast, is a mathematical formula. Figure 1 directly encodes an expression that can be written down, analysed, and interpreted. This distinction is important to us because we want an explicit equation that can be examined and potentially connected to underlying physics.

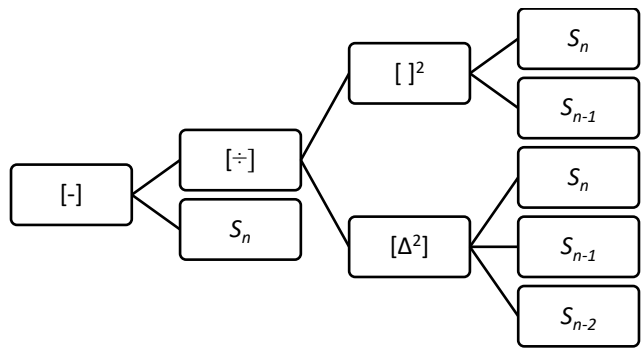

Figure 1: Tree representation of Aitken's delta-squared method. The terminals are sequence values; the functions are subtraction, division, square, and second difference.

**II.B. Evolution Mechanics**

GP evolves a population of candidate trees over successive generations [15]. The process mimics natural selection – trees that perform well are more likely to reproduce, passing their structure to the next generation of formulas. Over time, the population tends toward better solutions. The key concepts to understand here are selection, crossover, and mutation.

Selection determines which trees reproduce. Tournament selection [18] is a common approach where one randomly picks a small group of trees from the population, and the one with the best fitness wins the tournament and becomes a parent for the next generation. This provides natural selection pressure towards better solutions while maintaining diversity.

Crossover combines two parent trees to produce an offspring [18]. A random node is selected in each parent, and the subtrees rooted at those nodes are swapped. Figure 2 illustrates this process. In this example, the node [-] in parent A is selected along with the subtree containing Y and Z. Similarly, the node [÷] node in parent B is selected with its subtree containing X and 2. These two subtrees swap places producing two new offsprings. Offspring A inherits the main structure from parent A but now contains parent B's division subtree, while offspring B gets the reverse. Therefore, if one parent has discovered a useful ratio structure and another has discovered a useful difference structure, crossover might combine them into a single tree. This recombination of building blocks is the primary driver of GP's search capabilities.

Mutation introduces random changes to a single tree [18]. A node is selected at random, and the subtree rooted there is replaced with a newly generated random subtree. Mutation essentially injects structures into the population that might not arise from recombining trees. This helps population explore newer regions of the search space, potentially avoiding local optima. Figure 3 demonstrates this process. The node labelled [-] is randomly selected for mutation and the subtree associated with that node is replaced by a new one.

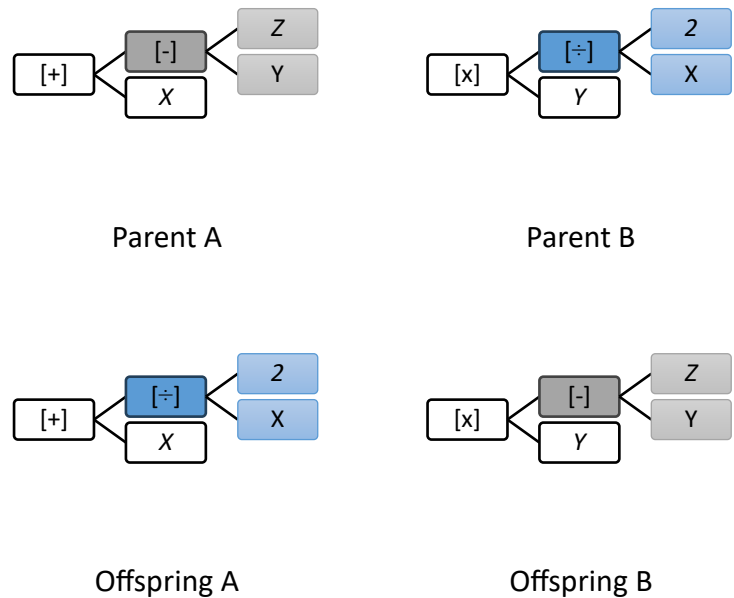

Figure 2: Crossover selects a node in each parent and swaps the subtrees (marked in grey) rooted in these nodes, producing new offsprings.

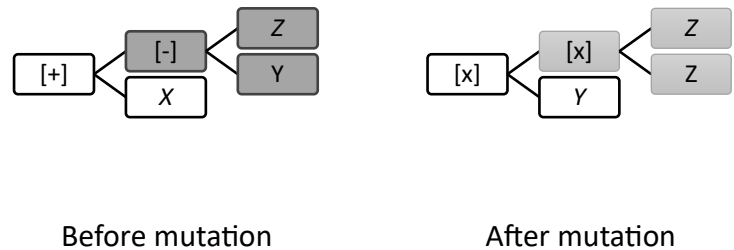

Figure 3: Mutation replaces a randomly selected subtree with a newly generated one.

This function drives the entire evolution process and pushes the population towards the appropriate solution via selection. We note that pure selection and variation can sometimes lose good solutions. In other words, a solution with high fitness score might fail to be selected, or lost to crossover, breaking useful structure. In order to preserve this structure, elitism is introduced. Elitism copies best trees directly into the next generation unchanged [15], which guarantees that the populations' best fitness does not degrade.

Having introduced the individual components – selection, crossover, mutation, and elitism, we can now investigate how they come together. Table 1 presents the complete evolution process. The algorithm begins by generating a random initial population of trees. Each tree is evaluated on the training data to determine its fitness. Evolution then proceeds for a fixed number of generations, or until some termination criterion is met. Within each generation, the algorithm first copies the best-performing trees directly into the new population (via elitism). The remaining slots are filled by selecting parents through tournaments, applying crossover, and mutation with their respective probabilities. The offspring are added to the new population which replaces the old one after all trees have been evaluated.

**Algorithm 1: Genetic Programming Evolution**

Input: Terminal set, Function set, Population size N,
    Max generations G, Crossover rate $p_c$, Mutation rate $p_m$,
    Elite count e, Training data

Output: Best tree observed

1.  Initialize population of N random trees
2.  Evaluate fitness of each tree on training data
3.
4.  FOR generation = 1 to G DO
5.     Create empty new population
6.
7.     // Elitism: preserve best trees
8.     Copy top e trees to new population unchanged
9.
10.    // Fill rest of population
11.    WHILE new population size < N DO
12.       // Selection: pick parents via tournament
13.       parent_A ← tournament_select
14.       parent_B ← tournament_select
15.
16.       // Crossover: combine parents
17.       IF random() < $p_c$ THEN
18.          (offspring_A, offspring_B) ← crossover(parent_A, parent_B)
19.       ELSE
20.          (offspring_A, offspring_B) ← (parent_A, parent_B)
21.       END IF
22.
23.       // Mutation: random modifications
24.       IF random() < $p_m$ THEN
25.          offspring_A ← mutate(offspring_A)
26.       END IF
27.       IF random() < $p_m$ THEN
28.          offspring_B ← mutate(offspring_B)

```
29.        END IF
30.
31.        Add offspring_A, offspring_B to new population
32.    END WHILE
33.
34.    Evaluate fitness of new trees
35.    population ← new population
36.
37.    IF termination criteria met THEN
38.        BREAK
39.    END IF
40. END FOR
41.
42. RETURN best tree from final population
```

Table 1: Pseudocode for genetic programming evolution.

Over successive generations, selection pressure favors trees with higher fitness, as crossover combines useful building blocks from different trees, and mutation introduces novel structures. The interplay of these building blocks drives the population towards better solutions. The specific parameter choices like population size, crossover and mutation rates, tournament size, and termination criteria are problem-dependent and are usually found using trial and error. Specifics for our transport problem at hand are detailed in the next section.

### III. GENETIC PROGRAMMING APPLIED TO TRANSPORT PROBLEMS

We introduced the general framework of genetic programming in the previous section. We will now describe its application to discovering convergence acceleration formulas for discrete ordinates transport. This section covers problem setup, GP configuration, and implementation details.

### III.A. Problem Setup

We consider the steady state, mono-energetic neutron transport equation in slab geometry with isotropic scattering [19]:

$$\mu \frac{\partial \psi(x,\mu)}{\partial x} + \Sigma_t \psi(x,\mu) = \frac{\Sigma_s}{2} \int_{-1}^{1} \psi(x,\mu') d\mu' + Q, \qquad (2)$$

where $\psi$ is the angular flux, $\mu$ is the direction cosine, $\Sigma_t$ is the total cross-section, $\Sigma_s$ is the scattering cross-section, and $Q$ is an internal source. We solve this equation using an analytical $S_N$ method [20] without spatial discretization error. Gauss-Legendre quadrature provides the angular discretization. This approach yields exact solutions for each quadrature order $N$, providing clean convergence sequences as $N$ increases.

To train and evaluate acceleration formulas, we generated 240 S_N solution sequences spanning a broad space of physical parameters. While scattering ratios c ranged from 0.001 to 0.999, the quadrature orders went from N=4 up to N=52 in increments of 4. For each parameter combination, the scalar flux was computed using an analytical S_N method [20], providing the desired sequences of center fluxes for relevant slabs. As $N$ increases, the solution converges to a continuous solution limit. The scattering ratios characterize how strong scatter is compared to absorption. Problems with scattering ratios limiting to unity often converge more slowly and could benefit from acceleration more.

We evaluate acceleration methods by comparing how quickly their outputs converge. For each method – raw, Aitken, Wynn-epsilon, and the evolved formula, we construct a sequence of values by applying relevant methods. We then compute the relative error between consecutive values:

$$\epsilon_n = \frac{|S_n - S_{n-1}|}{|S_n|}, \tag{3}$$

where $S_n$ is the sequence value with index $n$. An acceleration formula succeeds at a given index if its relative error is smaller than that of relative to the raw sequence.

$$\epsilon_n^{formula} < \epsilon_n^{raw}. \tag{4}$$

This criterion measures convergence rate rather than absolute accuracy. Smaller relative error indicates the sequence is stabilizing more quickly. We evaluate errors at positions $S_{20}$, $S_{28}$, $S_{36}$, $S_{44}$, and $S_{52}$, which span the later portion of the sequence where solutions set can benefit from convergence acceleration.

### III.B. Genetic Programming Configuration

Our terminal sets consist of sequence values $S_n$, $S_{n-1}$, $S_{n-2}$, and $S_{n-3}$, along with a learnable parameter $p$. The four consecutive sequence values provide raw material for acceleration. This window accommodates classical methods like Aitken (which considers the last three terms) and allows for discovery of formulas requiring additional terms. The learnable parameter enables fine-tuning of evolved formula if necessary. The function set includes binary, unary, and special operations. The binary set includes simple addition, subtraction, multiplication, and division. The unary set includes square, and the special set of operations includes second differences. Protected division returns a large value ($10^6$) when denominator falls below $10^{-10}$, ensuring closure. The second difference operator is included because it appears in Aitken's method and captures sequence curvature. We kept the function set small for this scoping study.

Transcendental functions could expand the search space but also increase effort and complexity. Therefore, they were not considered for this work.

Based on preliminary experiments, we used the following configuration for our dataset and problem at hand. We used a population size of 40 sample formulas for each generation. The maximum number of generations was set to 200 to avoid lengthy computations. The modest population keeps computational cost manageable while still maintaining genetic diversity. While the crossover rate was set to 70%, the mutation rate was set to 30%. The high crossover rate promotes recombination of useful building blocks, while mutation rate introduces novel structures. Two best formulas for each generation were automatically passed on to the next one without any changes by setting elite count to 2. Elitism ensures best solutions are not lost. To select parents for reproduction, we use tournament selection with tournament size 3. For each parent needed three individual formulas are randomly drawn from the population, and the one with the highest effectiveness wins the tournament and becomes the parent. The size of 3 provides moderate selection pressure and helps maintain population diversity while still driving improvement. The tree depth is limited to 4 to avoid overly complicated formulations. These parameters balance exploration and exploitation.

After each tree is created or modified, the learnable parameter is optimized to maximize the performance on training data. This hybrid approach of evolving structure while numerically optimizing parameters lets GP focus on discovering good functional forms while delegating coefficient tuning to numerical optimization. We use a 70/30 train/test split with 168 sequences being used for training and the remaining 72 being used for validation. The fitness function during evolution uses only training data and validation is performed separately to check for veracity of the formula obtained. Fitness is defined as the fraction of training cases where evolved formula beats raw sequence (or ones from other methods). With 168 training sequences and 5 evaluation points, there are 840 total comparisons per fitness evaluation. Our target rate was set to 75% which is roughly double of what classical methods achieved on this specific dataset.

### III.C. Implementation

Mathematical experiments were conducted using MATLAB 2025a [21] on a workstation with dual Intel Xeon Gold 6148 processors and 192 GB of RAM. The random number generator was seeded with a fixed value to ensure reproducibility. The evolved formulas took the form of a rational function where a numerator tree divided by a denominator tree to naturally accommodate the structure formulas like Aitken's method. To give the search a reasonable starting point, the initial formula set is seeded with a copy of Aitken's formula. The remaining formulas are generated randomly with varying tree depths to promote diversity.

During evolution, each candidate formula undergoes parameter optimization using MATLAB's fminsearch [21] with multiple random restarts to avoid local minima. The formula is then applied to training sequences, constructing a list of accelerated values at each evaluation position. Consecutive relative errors are computed and compared against those of the raw sequence. Formulas producing invalid outputs are penalized by counting those cases as losses. Evolution proceeds until either maximum iteration count is reached or the success rate exceeds 75%. The evolution typically takes roughly two hours to complete. The results presented in the next section were obtained from a run that achieved the 75% target at generation 63.

## IV. RESULTS AND DISCUSSION

This section presents results of applying GP to discover convergence acceleration formulas for $S_N$ solutions. We describe the discovered formula, evaluate its performance against classical methods, and discuss validation results and their implications.

### IV.A. Discovered Formula and Results

The GP search converged at generation 63, discovering the following formula:

$$A_n = \frac{S_n S_{n-2} - S_n^2 - S_{n-1}^2}{(2S_n - 4S_{n-1} + S_{n-2}) \cdot \left(\frac{S_n}{S_{n-1}}\right)}. \tag{5}$$

The evolved formula takes a rational form that differs slightly from classical accelerators. Unlike Aitken's method, which assumes geometric convergence, or Wynn-epsilon, which builds continued fraction approximations, the discovered formula incorporates cross-term products and second differences that appear to capture structure required to accelerate $S_N$ sequences.

The evolved formula achieved improved acceleration in more than 75% of test cases – approximately twice the success rate of classical methods on this dataset. Table 2 summarizes the performance comparison.

| Method | Success rate vs. Raw [%] |
|---|---|
| Aitken | 39 |
| Wynn-epsilon | 32 |
| GP-evolved | 78 |

Table 2: Success rates against unaccelerated sequences.

Figure 4 shows the performance evaluated by position in the sequence for the GP-evolved formula. The green regions indicate cases where acceleration improves convergence, while the red regions indicate cases where raw sequence converged faster (or had a better relative error). We observe that the performance of GP-evolved accelerator did not degrade with increasing scattering ratio or quadrature order, suggesting that the formula captured fundamental convergence behavior instead of fitting to regimes related to specific physical conditions.

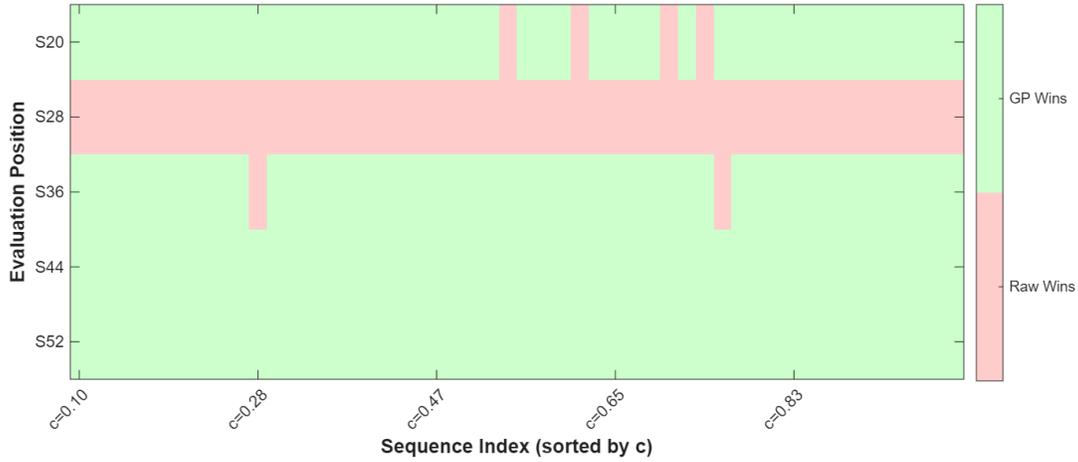

Figure 4: Heat map representing success rate for GP-evolved formula for different evaluation positions and scattering ratios.

To assess the generalization, we used a 70/30 train/test split. The formula evolved using the training set of 168 sequences and then evaluated on the held-out validation set of 72 sequences. The validation performance reached approximately 77.4%, with the gap compared to training performance of less than 0.5%. This minute degradation in performance suggests that the evolved formula generalizes well and does not overfit to the training data. The approach directly discovers formula structure rather than assuming a functional form and fitting parameters, which represents the novel approach of this study well.

**IV.B. Discussion**

The success of the GP-evolved formula demonstrates transport-specific acceleration patterns can be discovered algorithmically. Classical methods like Aitken and Wynn-epsilon were designed for general sequences with geometric or oscillatory convergence behavior in mind. Aitken's method is optimal for sequences converging geometrically. When this assumption holds, the method can dramatically accelerate convergence. However, $S_N$ sequences may not always exhibit this behavior depending on the regime the physical parameters push solutions into. The convergence rate can depend on the scattering ratio, the quadrature order, and the spatial location, producing patterns that vary across the parameter space [1]. Wynn-epsilon, while more general than Aitken, constructs approximations that may not always match the specific convergence structure of the $S_N$ solutions.

The GP approach makes no assumptions about convergence behavior. Rather, it learns structural features directly from the data. The cross-term products and second differences appearing in the discovered formula suggest sensitivity to how the convergence rate changes.

The nested second difference structure may act as a detector of convergence rate variation, adjusting acceleration dynamically based on local sequence behavior. This adaptability could explain why formula maintains performance across different scattering ratios and quadrature orders. The roughly double improvement over classical methods is notable. While classical methods have been refined over the years with careful mathematical analysis of convergence behavior, data-driven searches like the ones presented in this paper may serve as an alternative to discover novel, undiscovered patterns.

The small gap between training and validation performance deserves attention. Overfitting is a persistent concern in machine learning approaches [17]. The generalization we observe for this dataset could stem from two factors – a) our function set is kept small limiting overly complicated formulations, and b) approach discovers formula structure rather than fitting parameters to a fixed functional form. A formula that achieves high performance through structural properties rather than fine-tuned properties is more likely to generalize, since the structure captures underlying patterns rather than noise in training data.

We note several limitations of the current work. The formula evolved and tested on a single problem class. While the range of the scattering ratios was broad, other problem features like anisotropic scattering, nonuniform/multidimensional geometries, or energy dependence were not considered. Additionally, while the formula outperforms classical methods empirically, we do not have a complete theoretical understanding of why this structure works. Establishing such understanding would guide us to design even better accelerators. Having said that, despite the remaining questions about the theoretical underpinnings and broader applicability, the results presented here establish that GP can discover effective transport-specific formulas that outperform classical methods. We conclude this study in the next section.

## V. CONCLUSIONS

This work demonstrated that GP can discover novel convergence acceleration formulas for $S_N$ solution sequences. The evolved formula achieved a 78% success rate – approximately double the performance of classical methods like Aitken's delta-squared. Decent generalization was observed with validation performance within 0.5% of the training results. This suggests that the approach captures genuine convergence patterns rather than overfitting to training data.

The key contribution of this work is automated discovery of transport-specific acceleration structure. Rather than assuming a functional form and fitting parameters, GP searches the space of possible formulas directly. The resulting formula incorporates structure without making any assumptions purely based on the data it observes. This is distinct from how acceleration methods are traditionally designed. This work suggests that data-driven symbolic regression can complement traditional approaches in computational physics. When problem-specific structure exists but is theoretically difficult to derive, evolutionary searches could offer a path to discovery.

Several directions for future work emerge from this study. The functional set could be expanded to include transcendental functions like exponentials and logarithms, potentially enabling discovery of more sophisticated acceleration patterns. This would add more complexity to our approach and must be addressed carefully. Therefore, multi-objective evolution to balance accuracy and simplicity must be considered to yield more interpretable results. Theoretical analysis of why the discovered structure works would help understand the specifics of effectiveness and limitations more robustly and help guide design of more effective accelerators. Finally, the methodology could be used to address a more comprehensive set of transport problems that would include anisotropic scattering, multidimensional geometries with non-uniform properties, and energy dependence.

This work is dedicated to Prof. Ganapol, whose five decades of contributions to transport theory have shaped the field profoundly. His pioneering work on semi-analytical methods and benchmarking has provided the foundation upon which this study builds. We offer this work to him as a retirement present and hope this honors his legacy of inspiring continued exploration of experimental mathematics.